%% file: main.tex
\documentclass[10pt,twocolumn,letterpaper]{article}

\usepackage{iccv}      
\usepackage[toc,page]{appendix}

\input{preamble}

\definecolor{cvprblue}{rgb}{0.21,0.49,0.74}
\usepackage[pagebackref,breaklinks,colorlinks,allcolors=cvprblue]{hyperref}

\def\paperID{11053} 
\def\confName{ICCV}
\def\confYear{2025}

\title{\nameMethod: Towards 3D-Consistent Video Generators}

\author{
	Chun-Hao Huang$^{1}$\quad
	Niloy Mitra$^{1,2}$\quad
	Hyeonho Jeong$^{1,3}$\thanks{}\quad
	Jae Shin Yoon$^{1}$\quad
	Duygu Ceylan$^{1}$\\
	\textsuperscript{1}Adobe Research \quad
	\textsuperscript{2}University College London \quad
		\textsuperscript{3}KAIST \\
	\url{https://paulchhuang.github.io/jog3rwebsite} 
}
\input{macro}
\begin{document}
\twocolumn[{%
\renewcommand\twocolumn[1][]{#1}%
\maketitle
\begin{center}
    \centering

    \includegraphics[width=\linewidth]{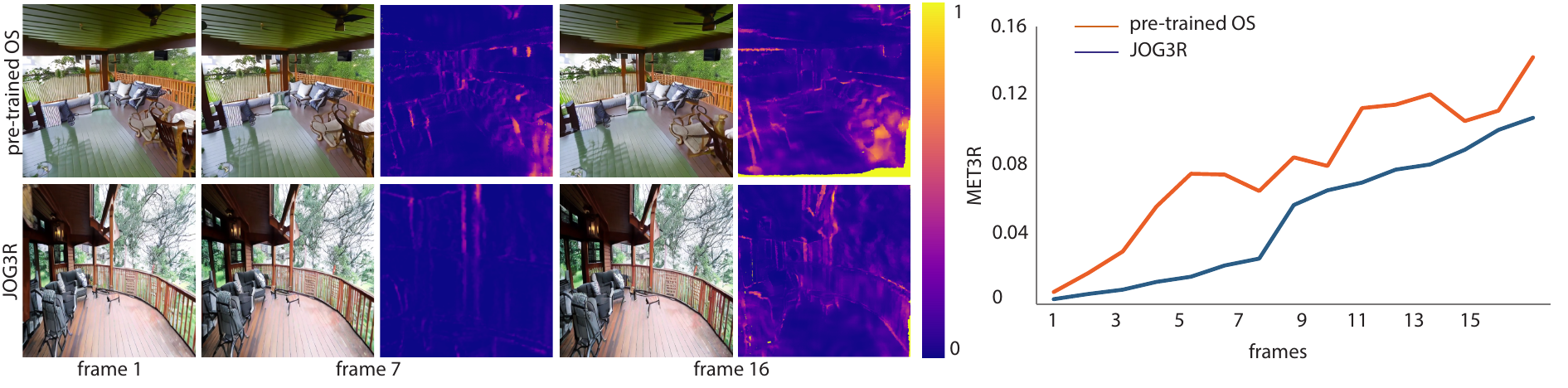}
    \vspace{-1em}
    \captionsetup{type=figure}
    \caption{We present \textbf{\nameMethod}, a unified framework that fine-tunes a video generation model \textit{jointly} with a 3D point map estimation task. \nameMethod improves the 3D-consistency of the generated videos compared to the pre-trained video diffusion transformer (DiT) as shown by the warped feature maps~(left) and scores~(right) using \meter~\cite{asim24met3r}, lower scores indicating higher 3D-consistency across frames.
    }
    \label{fig:teaser}
\end{center}%
}]
\input{sec/00_abstract}
\input{sec/01_intro}

\input{sec/02_background}

\input{sec/03_methodology}
\input{sec/04_experiments}

\input{sec/05_limitation_conclusion}
{
    \small
    \bibliographystyle{ieeenat_fullname}

\input{main.bbl}
}

\clearpage
\begin{appendices}
	\input{sec/10_suppmat}

\end{appendices}

\end{document}


\maketitlesupplementary
 
The supplementary material of submission \textit{11053} consists of this document and the webpage. We provide qualitative results in our webpage.

\section{Temporal Regularizer Terms}
Given the estimated camera pose $(\textbf{R}_f,\textbf{t}_f)$ and focal length $l_f$ for each frame $f$, we define
$\flcloss$ and $\temploss$ as below:
\begin{align}
\flcloss &= \sum_f \left \| l_f - l_{f+1}\right \|_2^2 \\
\temploss &= \sum_f \left \| \textbf{t}_f - \textbf{t}_{f+1}\right \|_2^2 + \left \| \textbf{v}_f - \textbf{v}_{f+1}\right \|_2^2
\end{align}
where $\textbf{v}_f = \textbf{t}_f - \textbf{t}_{f+1}$. The first term of $\temploss$ encourages static camera position while the second term encourages constant velocity.

\section{Architectural Design Choices}
In Table 1 below, we compare the quality of estimated camera poses using the feature maps in different DiT block $b^i$. We observe that earlier blocks ($i$=10) yields lower errors in RealEstate10k-test but higher errors in DL3DV10k, suggesting that using features of earlier blocks has a higher risk of poor generalization. We conclude that features of later blocks, e.g., $i\in [20, 27]$ are preferred than earlier blocks. 
Instead of solely relying on one block $b^i$, one can potentially devise a module fusing features of all DiT blocks and projecting to the input space of \duster decoders, which we consider future work.


\begin{table*}[b]
\begin{center}
\footnotesize
\begin{tabular}{lrrrrr}
\toprule
$b^i$  & Rot.~err.~($^\circ$) $\downarrow$ &  Transl.~err.~($\degree$) $\downarrow$ & RRA@5$\degree$ $\uparrow$ &  RTA@5$\degree$  $\uparrow$  & mAA@30$\degree$ $\uparrow$ \\
\midrule
& & & RealEstate10k \\
\midrule

$b^{10}$ & 0.28 & 18.94 & 99.94\% & 18.75\% & 52.63\% \\
$b^{20}$ & 0.28 & 19.10 & 99.85\% & 21.59\% & 53.35\% \\
$b^{26}$ (\nameMethod) &  0.29	& 22.15 & 99.79\% & 19.18\% &	47.25\%  \\
$b^{27}$ & 0.28	& 23.09 & 99.80\% & 17.74\% &	46.88\%  \\
\midrule
& & & DL3DV10k \\
\midrule
$b^{10}$ & 4.68 & 30.90 & 73.87\% & 6.08\% & 30.76\% \\
$b^{20}$ & 4.01 & 27.71 & 77.91\% & 10.56\% & 36.62\%\\
$b^{26}$ (\nameMethod) & 4.20	& 29.17 & 78.63\%& 7.86\% & 34.22\% \\
$b^{27}$ & 4.65	& 30.36 & 77.15\% & 7.98\% &	32.86\%  \\
 \bottomrule
 \end{tabular}
\end{center}
\vspace{-2mm}
  \caption{
 \textbf{Ablation study on which feature maps $b^i$ get passed to \duster.}
 }
 \label{tab:ablation} 
\end{table*}


%% file: preamble.tex
\usepackage{gensymb}


%% file: sec/00_abstract.tex
\begin{abstract}

Emergent capabilities of image generators have led to many impactful zero- or few-shot applications. Inspired by this success, we investigate whether video generators similarly exhibit 3D-awareness. 
Using structure-from-motion as a 3D-aware task, we test if intermediate features of a video generator -- OpenSora in our case -- can support camera pose estimation. Surprisingly, at first, we only find a weak correlation between the two tasks. Deeper investigation reveals that although the video generator produces plausible video frames, the frames themselves are not truly 3D-consistent. 
Instead, we propose to jointly train for the two tasks, using photometric generation and 3D aware errors. Specifically, we find that SoTA video generation and camera pose estimation (i.e., \duster~\citep{dust3r_cvpr24}) networks share common structures, and propose an architecture that unifies the two. 
The proposed unified model, named \nameMethod, produces camera pose estimates with competitive quality while producing 3D-consistent videos. 
In summary, we propose the first unified video generator that is  
3D-consistent, generates realistic video frames, and can potentially be repurposed for other 3D-aware tasks.

\end{abstract}

%% file: sec/01_intro.tex
\section{Introduction}

Following the success of foundational image generators~\cite{rombach2021highresolution}, video generators have quickly become a reality. After the initial demonstration by Sora, many similar models have rapidly emerged, both in commercial and open-source domains~\citep{guo2023animatediff,opensora,videoworldsimulators2024,menapace2024snap,blattmann2023stable}. Trained on large-scale datasets (e.g., WebVid-10M \citep{Bain21}, Panda-70M \citep{panda70m}), these models produce impressive diversity with both compelling image quality and temporal consistency. Given the emerging behaviors observed in the case of image generators~\cite{Position:Platonic} leading to several successful zero- or few-shot approaches (e.g., feature detection\cite{tang2023emergent,dutt2024diffusion}, segmentation~\cite{ni2023refdiffzeroshotreferringimage}, generative editing~\cite{pandey2024diffusionhandles}), we investigate \textit{if video generators can be similarly repurposed for 3D-aware tasks} (e.g., structure from motion, camera pose, correspondence tracking).

As a 3D-aware task, we pick the classical structure-from-motion~(SfM) problem as it requires reasoning about both scene geometry and the relative viewpoint changes across frames. We expect this 3D task to be compatible with video generation as both tasks need to arrive at feature representations that capture the physical world~\cite{Huh24}. We further observe that the ViT backbone of the \duster architecture~\cite{dust3r_cvpr24}, which is the leading feedforward backbone for establishing correspondence between video frames, actually shares many architectural designs with the Diffusion Transformers (DiT) in state-of-the-art video generators. This allows us to \emph{stitch} the OpenSora backbone with a \duster-like point map -- and hence camera pose -- estimation head into a unified architecture. 

We train the 3D point map reconstruction head using a frozen video generator (we used OpenSora~\cite{opensora} in our tests). Somewhat surprisingly, we find although the chosen tasks are seemingly compatible, the estimated 3D point maps and relative camera poses were, at best, mediocre (see Section~\ref{sec:exp}). On deeper investigation, we found that although the (pre-trained) video generators produced visually compelling and temporally-smooth frames (as indicated by FVD scores), they were \textit{not 3D-consistent}. This is also exposed by computing warped feature scores using \meter~\cite{asim24met3r} (see Figure~\ref{fig:teaser}); this explains why taking existing internal video generator features leaves a gap between the tasks of video generation and 3D estimation. 

The above insight helps us design a video generator that is both visually compelling (i.e., good FVD score) and 3D-consistent (i.e., good \meter score). 
In particular, we fine-tune the video generator using the proposed unified architecture supervised by both video generation and 3D geometric  (correspondence) losses. Such a formulation makes the two tasks more `equivalent', and hence improves both. Thus, the secondary task of 3D reconstruction helps improve the 3D coherence of the video generator (see Figure~\ref{fig:teaser} and the supplementary webpage for more results).

To summarize, we present a novel architecture unifying video generation with 3D point map estimation, which we refer to as \textit{JOint Generation and 3D Reconstruction}, in short \nameMethod. 
Being a unified model, \nameMethod generates videos (T2V), estimates 3D point maps, hence camera poses, given a video (V2C), or do both in one go (T2V+C). 
In particular, we test whether with fine-tuning one can produce video generator features that can also be reused for improved 3D reconstruction. Finally, we analyze the effect of such fine-tuning on generation quality. Our experiments show that while video generation features exhibit some 3D awareness natively (\ie, obtained from pre-trained generator directly), adapting them with additional supervision on the 3D reconstruction task boosts their 3D-consistency. As a side benefit, the additional supervision produces competitive camera pose estimations compared to state-of-the-art solutions, and unifies the video generation and camera pose estimation tasks. \textit{Code will be released on acceptance.}

%% file: sec/02_background.tex
\begin{figure*}[h]
\begin{center}
\includegraphics[width=\linewidth]{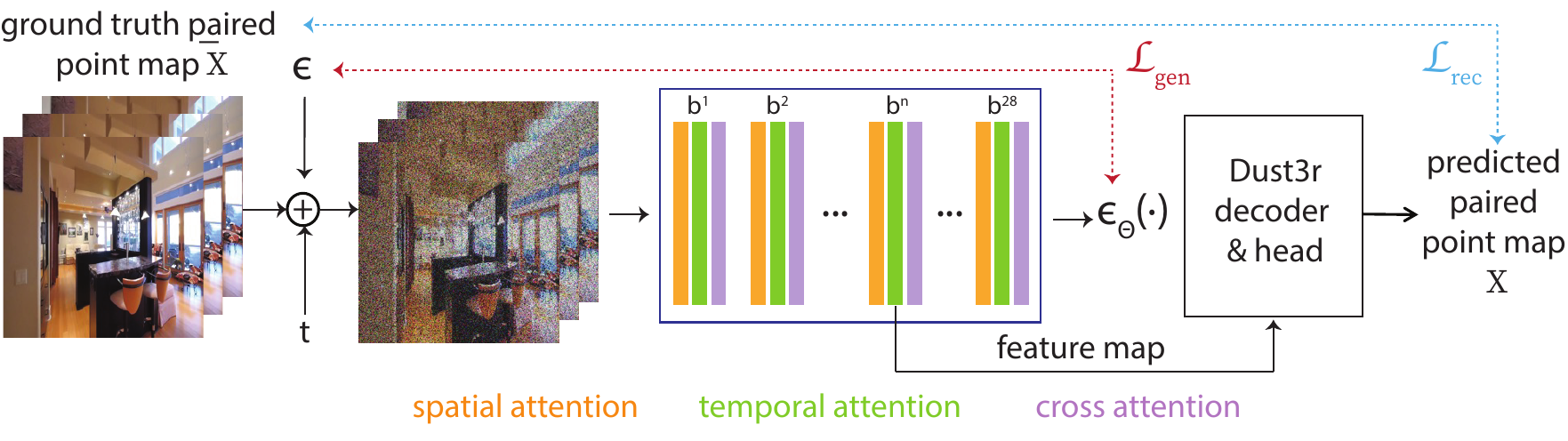}
\end{center}
\vspace{-2mm}
\caption{We propose a unified framework to investigate if the intermediate features from a video generation model can be repurposed for 3D point map estimation by routing them to the SoTA decoder of \duster. We investigate the effect of freezing vs training certain modules of the generator using different combination of generation and reconstruction losses.}
\label{fig:training}
\end{figure*}
\section{Related Work}
\subsection{Diffusion-Based Video Generation}
Building on the success of diffusion models \citep{ho2020denoising, song2020score} in image synthesis \citep{dhariwal2021diffusion, rombach2021highresolution}, the research community has extended diffusion-based methods to video generation.
Early works \citep{ho2022imagen,ho2022video} adapted image diffusion architectures by incorporating a temporal dimension, enabling the model to be trained on both image and video data.
Typically, UNet-based architectures incorporate temporal attention blocks after spatial attention blocks, and 2D convolution layers are expanded to 3D convolution layers by altering kernels~\citep{ho2022video, wu2023tune}.
Latent video diffusion models \citep{blattmann2023align, he2022latent, wang2023modelscope, blattmann2023stable} have been introduced to avoid excessive computing demands, implementing the diffusion process in a lower-dimensional latent space.
Seeking to generate spatially and temporally high-resolution videos, another line of research adopts cascaded pipelines \citep{ho2022imagen, singer2022make, zhang2023show, wang2023lavie, bar2024lumiere}, incorporating low-resolution keyframe generation, frame interpolation, and super-resolution modules.
To maximize computational scalability, recent waves in video generation \citep{chen2023gentron, ma2024latte, menapace2024snap, videoworldsimulators2024, opensora} diverge from UNet-based architecture and employ the Diffusion Transformer (DiT) \citep{peebles2023scalable} backbone that processes space-time patches of video and image latent codes.
Following this direction, we build our method on OpenSora \citep{opensora}, a publicly available DiT-based latent video diffusion model. These models are trained with only photometric error and, as demonstrated in our evaluation, not 3D-consistent (see Section~\ref{sec:exp} and \cite{asim24met3r}). This, in turn, means that their internal features are not suitable for mixing~\cite{modelMixing:15}, with no or little fine-tuning, for 3D-aware tasks~\cite{bahmani2025ac3d,jeong2024track4gen}. 
\subsection{3D Reconstruction}
The fundamental principles of multiview geometry~\cite{Wrobel2001MultipleVG} including feature extraction~\cite{Lowe2004DistinctiveIF,Brown2011DiscriminativeLO}, matching~\cite{Agarwal2009BuildingRI,Lou2012MatchMinerES,Wu2013TowardsLI,Havlena2014VocMatchEM}, and triangulation with epipolar constraints are well known to produce  accurate (yet sparse) 3D point clouds with precise camera pose estimation from multiview images of real scenes~\cite{schonberger2016structure}.
The efficiency of 3D reconstruction has been improved with linear-time incremental structure-from-motion~\cite{Wu2013TowardsLI}
and coarse-to-fine hybrid approaches~\cite{crandall2012sfm,cui2017hsfm}.
To improve robustness to outliers, researchers proposed global camera rotation averaging~\cite{cui2017hsfm}, camera optimization techniques based on features of points vanishing with oriented planes~\cite{holynski2020reducing}, or from a learned neural network~\cite{lindenberger2021pixel} to prevent rotation and scale drift issues.
%
%
Global camera pose registration and approximation with geometric linearity~\cite{jiang2013global,cai2021pose} or joint 3D point position estimation~\cite{pan2024glomap} are designed to further push the scalability and efficiency of the 3D reconstruction as well as the robustness particularly to the image sequence with small baselines.


Given estimated camera poses and sparse 3D point clouds, multiview stereo can then produce a dense 3D surface using hand-created visual features~\cite{schonberger2016pixelwise} or neural features with a cost volume~\cite{ma2022multiview,ummenhofer2021adaptive,ma2022multiview,zhang2023geomvsnet,ye2023constraining} to predict globally coherent depth estimates.
Existing neural rendering methods reconstruct such a dense surface by modeling the implicit or explicit cost volume
and differentiable rendering of the scene for photometric supervision from multiview images~\cite{li2023neuralangelo,sun2022neural,peng2023gens,Guo2022Neural3S,Yu2022MonoSDFEM,Wang2022ImprovedSR,Oechsle2021UNISURFUN,Wang2021NeuSLN,Murez2020AtlasE3} or monocular depth estimation~\cite{Sayed2022SimpleRecon3R}. 
Some pose-free methods further erase the requirement of camera calibration: test time optimization produces globally consistent depth map under unknown scale and poses using frozen depth prediction model~\cite{xu2023frozenrecon}; the unsupervised signals from dense correspondences such as optical flow are integrated to learn from unlabeled data~\cite{Yin2018GeoNetUL,Teed2018DeepV2DVT,Zhou2019DeepTAMDT}.
Recent works proposed a direct regression framework for dense surface reconstruction from pairwise images by learning to predict globally coherent depths and camera parameters~\cite{Ummenhofer2016DeMoNDA} or to directly predict per-pixel 3D point clouds from two views~\cite{dust3r_cvpr24,mast3r_arxiv24} using a vision transformer with dense tokenization~\cite{Ranftl2021VisionTF}. 
However, these methods are designed for real videos, and fail to handle generated videos, which so far have \textit{not} been 3D-consistent. This has also been observed in recent warped feature analysis among video frames as reported by \meter~\cite{asim24met3r}.

\subsection{Diffusion Model as Features for Reconstruction}
A generative diffusion model is often trained on millions of paired image and text prompts and in the process develops a semantically meaningful visual prior. 
Naturally, researchers are interested in using this strong prior for many downstream 3D vision tasks.
Injecting 3D awareness into the diffusion prior greatly improves the accuracy and generalizability of the monocular depth estimation and correspondence search tasks~\cite{prob3d,yue2024improving}.
The latent features from the frozen pretrained diffusion model are often used as a backbone, and a task-specific decoder with cross attention is newly trained for semantic correspondences~\cite{tang2023emergent,zhang2023a,hedlin2024unsupervised,zhang2024telling,hedlin2024unsupervised,jiang2024omniglue,dutt2024diffusion}, semantic segmentation and monocular depth estimation~\cite{zhao2023unleashing}, material and shadow prediction~\cite{zhan2023does}, general object 3D pose estimation~\cite{ornek2023foundpose,cai2024open}.
However, such image diffusion features do not inherently consider the temporal relation between the frames, leading to temporally unstable 3D prediction results from videos.
In contrast, we investigate video diffusion features as a backbone for the multitasking prediction of video generation and 3D camera poses estimation (see \cite{Position:Platonic,bahmani2025ac3d}). Recently, video generators have been probed for multiview consistency --- \meter evaluates multiview consistency by performing dense 3D reconstructions from image pairs using \duster~\cite{dust3r_cvpr24}, and uses the estimated tranformation to warp image contents from one view into the other. The warped features are then compared, using a view-independent similarity score, with lower scores indicating higher multiview consistency -- for static scenes, this measures quality of 3D-consistency~(see also, Figure~\ref{fig:teaser}). 







%% file: sec/03_methodology.tex
\section{Method}
\subsection{Preliminaries}

\minisection{Video diffusion model.} We consider \opensora \citep{opensora} as our base video generation model, which is a \dit-based video diffusion model inspired by the notable success of Sora \citep{videoworldsimulators2024}. 
It performs the diffusion process in a lower-dimensional latent space defined by a pre-trained VAE encoder $\mathcal{E}$. 
Each frame $x$ of the input video is first projected into this latent space, $\latent_0=\mathcal{E}(x)$.
Given a diffusion time step $\timestep$, the \textit{forward} process incrementally adds Gaussian noise to the latent code $\latent_0$ via a Markov chain to obtain noisy latent $\latent_{\timestep}$. 
The denoising model $\epsilon_\theta$ takes the noisy latents of all frames, the time step $\timestep$, and the text prompt $\prompt$ as input to predict the added noise: $\epsilon_\theta ( \{\latent_{\timestep}^f\}_{f=1}^{F}, \timestep, \prompt)$, where $F$ is the total number of frames and $\theta$ denotes the parameters of the \dit network~\citep{peebles2023scalable}. The network consists of $28$ spatial-temporal diffusion transformer (ST\dit) blocks $\{b^1,\dots,b^{28}\}$, similar to \cite{ma2024latte}. 
The iterative process of noise prediction and noise removal is referred to as the \textit{reverse} process.

\minisection{3D reconstruction model.} We consider the state-of-the-art multi-view stereo reconstruction (MVS) framework \duster~\citep{dust3r_cvpr24} as our 3D module.
Given an image pair, \duster encodes each image independently with a ViT encoder \citep{dosovitskiy2020vit, weinzaepfel2022croco}.
Two decoders process both features to enable cross-view information sharing, followed by separate heads that estimate point maps $\pointmap \in \mathbb{R}^{H\times W\times3}$, represented in the coordinates of the first view as $\pointmaponeone$ and $\pointmaptwoone$, respectively. 
The relative camera pose is then estimated by aligning $\pointmaponeone$ and $\pointmaponetwo$ 
using Procrustes alignment \citep{procrustes_alignment} with PnP-RANSAC \citep{pnp,ransac}.
A global optimization scheme is employed to register more than a pair of views from the same scene as post-processing.

\begin{figure*}[h]
\begin{center}
\includegraphics[trim={0.9cm 0 0 0},clip,width=\linewidth]{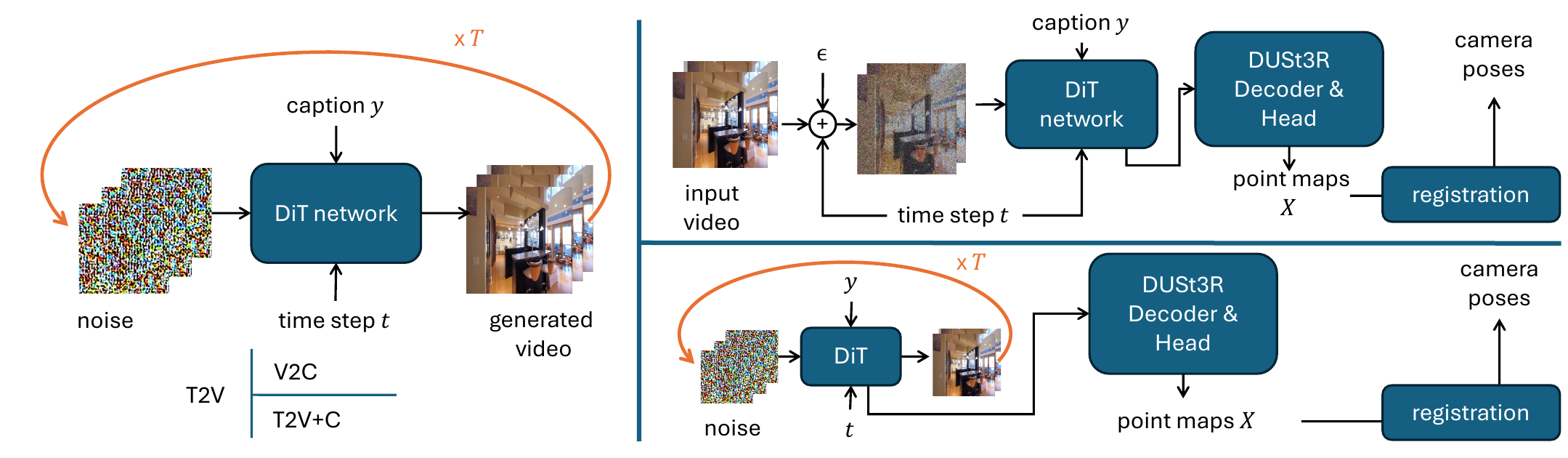}
\end{center}
\vspace{-7mm}
\caption{We base our analysis on three main tasks: text-to-video (T2V), video to camera estimation (V2C), and joint video generation and camera estimation (T2V+C) at inference time.}
\label{fig:inference}
\vspace{-2mm}
\end{figure*}

\subsection{Unified Video Generation \& 3D Reconstruction}
\minisection{Feature routing via model stitching.} We argue that both a foundational video generation model as well as a 3D reconstruction model are trying to arrive at feature representations that capture the physical reality and hence are likely to converge to feature spaces~\cite{Position:Platonic} that exhibit equivalence. To verify this hypothesis and facilitate our analysis, inspired by the \emph{model stitching} technique~\cite{lenc2015}, we propose a unified framework that routes intermediate features from the video generator to the 3D reconstruction task.
We observe that ViT and \dit share many architectural designs in common since they both belong to the broad transformer family.
Hence, our key insight is to replace the \textit{image}-based ViT encoder in \duster with the \textit{video} \dit backbone in \opensora. 
In other words, we stitch the transformer blocks of the \dit network $\epsilon_\theta$ with the \duster decoders and heads (see Figure \ref{fig:training} for illustration). In this setup, the intermediate features from the video generation model are effectively routed to \duster decoders to perform the reconstruction task. 
Specifically, 
we extract the output of the intermediate STDiT block $b^n$ at a particular time step $\timestep$ during the reverse process.
Following Tang \etal~\cite{tang2023emergent}, we consider small $\timestep$ where the feature focuses more on low-level details, making it useful as a geometric feature descriptor to build correspondence across frames.

\minisection{3D-aware training for video generation.} Given a stitched and unified model, as described above, a naive baseline is to use pre-trained features from the video model and only train the stitched \duster decoders to perform a 3D reconstruction task. As discussed in Section~\ref{sec:exp} and shown in Table~\ref{tab:rec_quanti}, while this baseline (row 1) provides reasonable 3D reconstruction quality, i.e., more effective than \duster trained on the same dataset (row 4) but not as effective as the pre-trained \duster (row 6). This shows that the feature representations of a video generator and a 3D reconstructor are compatible but the features of a pre-trained video generator are not fully 3D-consistent. This hinders performance. Hence, we further fine-tune both the video generator and the reconstruction heads jointly to perform generation and reconstruction tasks. 

For training, we consider two losses: generation loss $\genloss$ and reconstruction loss $\recloss$. The generation loss $\genloss$ is the common objective in training diffusion models that aims to match the added noise $\epsilon$ and helps to retain the generative power of the video model.
The reconstruction loss $\recloss$ is aimed to improve the 3D-awareness of the intermediate features and follows the formulation in \duster. It is defined as the sum of confidence-weighted Euclidean distance $L_2(f,i)$ between the regressed point maps $\pointmap$ and the ground truth point maps $\bar{\pointmap}$ over all valid pixels $i$ and all frames $f$. Formally, \begin{eqnarray}
\genloss &:=& \left \| \epsilon - \epsilon_\theta\left(\{\latent_{\timestep}^f\}_{f=1}^{F}, \timestep, \prompt \right)  \right \|_2^2
\\ 
\recloss &:=& \sum_{f \in \{ 2,\dots F\}} \sum_i C_i^{f\rightarrow 1} L_2(f,i) - \alpha \log C_i^{f,1} \\
&& \text{with} \;
L_2(f,i) = \left \| \frac{1}{s} \pointmap_i^{f\rightarrow 1} -  \frac{1}{\bar{s}} \bar{\pointmap}_i^{f\rightarrow  1} \right \|_2 \nonumber
\label{eq:recloss}
\end{eqnarray}
where the scaling factors $s$ and $\bar{s}$ handles the scale ambiguity between prediction and ground-truth by bringing them to a normalized scale, $ C_i^{f\rightarrow  1}$ is the confidence score for pixel $i$, which encourages network to extrapolate in harder areas, and $\alpha$ is a hyper-parameter controlling the regularization term \citep{Confnet} (see \cite{dust3r_cvpr24} for more details). In our experiments, we study the effect of different combinations of these two losses. Specifically, activating $\recloss$ alone analyzes the native 3D awareness of the video generation features while using them in conjunction, $\mathcal{L}_\text{total} =  \genloss + \lambda \recloss$ (we empirically set $\lambda=1$) investigates if the features can further be adapted for both video generation and camera pose estimation tasks. \nameMethod fine-tunes the video model starting with its pre-trained weights while the \duster decoder and heads stitched to this generation model are always trained from scratch.

\minisection{Our modification of \duster.} The features extracted from the video generator encode a sequence of $F$ frames and are provided to \duster in a pair-wise manner.
During training, the first frame can ideally be paired with all other frames $f$. In practice, due to memory constraints, we sample 4 pairs from the set $\{(1\rightarrow f)\}_{f=2}^F$ to predict the 3D point maps.

Once the unified model is trained, if it is desired to obtain camera parameters for a given sequence, at inference time, we first predict the point maps between all pairs $(1\rightarrow f)$ and perform the global camera registration in \duster to refine the camera pose, depth and focal length for each frame, akin to bundle adjustment.
Since our input is not a set of sparse views but a temporal sequence, we append two regularization terms, $\temploss$ and $\flcloss$, to the original global registration objective, encouraging smooth camera translation and consistent focal length between neighboring frames, respectively. 
The two terms are used only at inference, and we refer to the supplemental for the detailed formulation.

%

\minisection{Implementation details.}
We adopt OpenSora 1.0 as our video generator, which uses 2D VAE (from Stability-AI)~\cite{rombach2021highresolution}, T5 text encoder \citep{raffel2020exploring}, and an STDiT (ST stands for spatial-temporal) architecture similar to variant 3 in \cite{ma2024latte} as the denoising network. 
Among the 28 STDiT blocks, we empirically set the first 4 frozen and update only the weights of the temporal attention layers for the remaining 24 blocks. We extract the output of the 26\textsuperscript{th} block $b^{26}$ as feature maps for \duster decoders. The final two blocks behave as a ``generation'' branch whose weights are only updated by the gradient of generation loss $\genloss$. We refer to the supplementary material where we describe how the features obtained from different STDiT blocks perform.

We adopt the linear prediction head of \duster for final point map estimation. \duster originally uses a decoder with 12 transformer blocks that is duplicated for each of the pair of frames. However, information sharing is enabled between the two decoders. In our experiments, we find that a decoder structure with six transformer blocks provides similar performance and report our results accordingly. Note that the decoder blocks used in our training experiments are trained from scratch. Furthermore, since the features we get from the generator encode all the frames in a video sequence, we also experiment with replacing the duplicate decoder architecture with a single decoder consisting of 6 transformer blocks that perform full 3D attention across all the frames. We empirically find that this performs on par with duplicate decoders (see Table~\ref{tab:rec_quanti}), and hence we use the latter to provide a more fair comparison to \duster. 

When training, we sample the time step $\timestep \in [0, 10]$ (corresponding to 10\% of noise level) and consider the empty prompt for computing the reconstruction loss $\recloss$, while for the generation loss $\genloss$ we sample the full range of time steps and use the captions of the videos as text prompts. 

%% file: sec/04_experiments.tex
\section{Experiments}
\label{sec:exp}
We base our analysis on evaluations conducted on three specific tasks (see Figure~\ref{fig:inference}). (i)~\textit{Text-to-video (T2V)}: We evaluate the effect of using additional supervision from the 3D reconstruction task on generation quality for the task of text to video generation where we sample Gaussian noise and iteratively denoise it with the text guidance. (ii)~\textit{Video-to-camera (V2C)}: We add noise to a given input video based on a sampled time step $\timestep$ sampled in the range $[0, 5]$, denoise it for one time step, route the feature maps to \duster decoders and heads, followed by registration of point maps $\pointmap$ to obtain camera poses. (iii)~\textit{Text-to-Video+Camera~(T2V+C)}: Once trained with a combination of generation and reconstruction losses, \nameMethod performs text-to-video generation while simultaneously routing the intermediate features to the reconstruction module at the desired time step (sampled in the range $[0, 5]$), without the overhead of adding noise and passing it through the network again. As a result, cameras are generated alongside the video \textit{in one go}, unifying the two tasks. 

We follow standard metrics to assess the generated video quality (FID and FVD for image/video quality; \meter for 3D-consistency) for T2V setup; while validating the accuracy of 3D reconstruction, hence camera pose estimation (for static videos), on real videos (V2C). We further provide results for jointly generating videos along with camera pose estimation (T2V+C). Since there is no ground truth in this case, we report self-consistency.

\subsection{Setup}
\minisection{Data.} 
We choose \realestate \cite{zhou2018stereo} as our main dataset, which has around 65K video clips of static scenes paired with camera parameter annotations. 
We use the captions of \realestate provided in \cite{cameractrl} and also follow their train/test split.
As pre-processing, we pre-compute the VAE latents of the video frames and the T5 text embeddings of the captions. We sample $F$=16 frames from the original sequences with a frame stride randomly chosen from $\{1,2,4,8\}$ and also randomly reverse the frame order with a probability of 0.5.

To obtain point map annotations $\bar{\pointmap}$, we estimate metric depth with ZoeDepth \citep{bhat2023zoedepth}, unproject them to 3D and transform to the coordinate of the first frame using the camera parameters in \realestate.
All camera extrinsic parameters are expressed with respect to the first frame.

In addition, to check generalization, we consider DL3DV10K \citep{ling2024dl3dv}, which also provides camera annotations, as an additional test set. We choose a random set of 70 videos for testing and caption the first frame of each video using \cite{li-etal-2023-lavis}. We prepare point map annotations using ZoeDepth \citep{bhat2023zoedepth}. 

\minisection{Baselines.}
Since there is no existing method that can perform both video generation and 3D reconstruction jointly, we can only compare to task-specific methods. For video generation, we consider the pre-trained OpenSora (\emph{pre-trained OS}) as well as a version fine-tuned on our dataset using generation loss only (\emph{fine-tuned OS}). We also define a baseline where we route the features from the pre-trained and frozen OpenSora model and only train the reconstruction heads using the reconstruction loss only (\emph{frozen OS + rec}). 
We use the original pair-wise method \duster~\cite{dust3r_cvpr24} with linear head 
as a camera pose estimation method to provide a reference for our analysis.
For \duster we consider three variants: (i)~off-the-shelf pretrained weights (DUSt3R\textsuperscript{$\dagger$}), (ii)~initialized with pretrained weights and trained with the same data as ours (DUSt3R*), and (iii)~trained from scratch with the same data as ours (DUSt3R$^0$). 
In all three variants, we perform the final global optimization step with our newly introduced temporal loss $\temploss$ and $\flcloss$.  

\begin{table}
\begin{center}
\footnotesize
\begin{tabular}{rccrrrr}
\toprule
Method & $\genloss$ & $\recloss$& FID $\downarrow$ & FVD $\downarrow$ & MEt3R  $\downarrow$ & \\
\midrule  \midrule
pre-trained OS & n.a. & n.a. & 115.36 & 1872.41 & 0.0819 \\
\midrule
frozen OS + rec  & \xmark & \cmark & 94.52 & 1797.07 &  0.0772 \\
fine-tuned OS  & \cmark & \xmark & 88.02 & \textbf{1440.92} &  0.0913\\
\nameMethod (ours)  & \cmark & \cmark & \textbf{79.94}	&  1742.73 & \textbf{0.0736} \\
 \bottomrule
 \end{tabular}
\end{center}
\vspace{-2mm}
  \caption{
 \textbf{Generation quality comparison.} We compute FID and FVD for photometric quality and \meter~\cite{asim24met3r} for 3D-consistency measure, both on the  RealEstate10K-test data. While being comparable in terms of video quality against OpenSora~(OS) baseline and variants, \nameMethod produces most 3D-consistent results. }
 \label{tab:gen_quanti}
 \vspace{-2mm}
\end{table}

\begin{figure*}[h]
\begin{center}
\includegraphics[width=\linewidth]{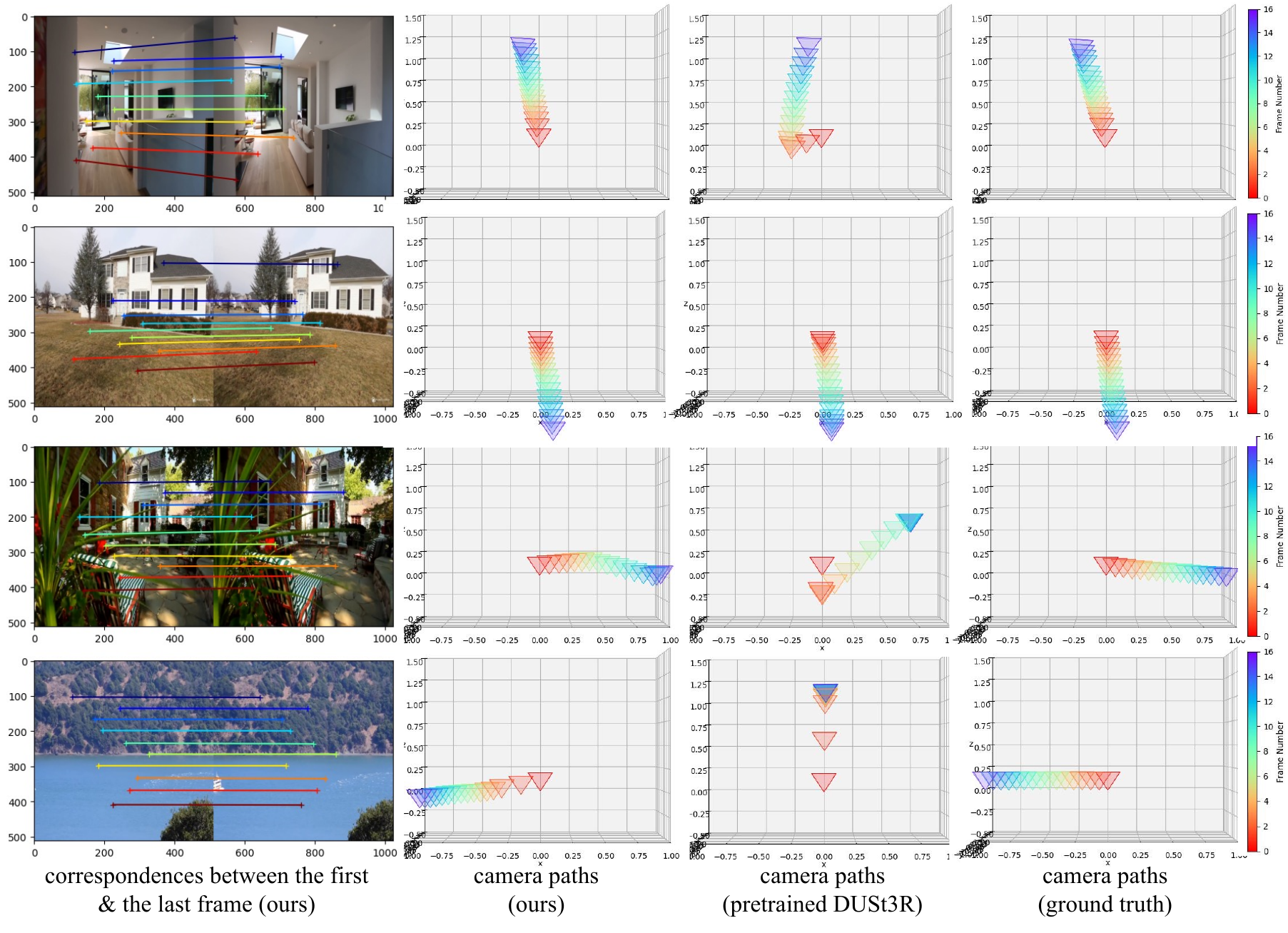}
\end{center}
\vspace{-2mm}
\caption{\textbf{Qualitative camera pose estimation (V2C) results.} Red to purple indicates the progression from the first to the last frame. Note that on these test videos, \nameMethod yields improved point maps leading to improved camera tracks compared to pretrained \duster. 
}
\label{fig:rec-visuals}
\end{figure*}

\begin{figure*}[h]
\centering
\includegraphics[width=.85\linewidth]{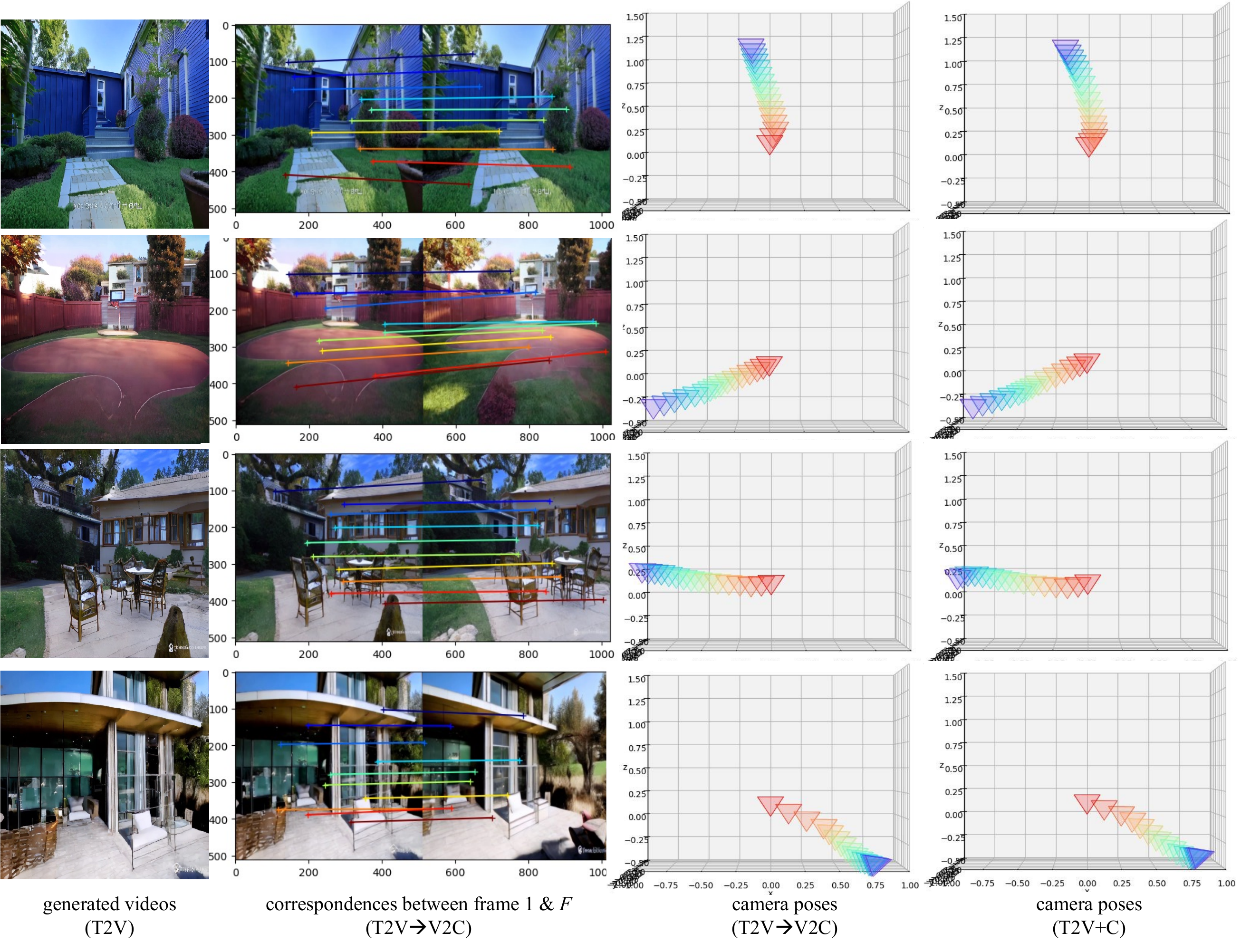}
\vspace{-5mm}
\caption{\textbf{Qualitative generation T2V+C results.} It is coherent with the camera paths from T2V$\rightarrow$V2C. Please see suppmat.~for videos.}
\label{fig:t2vc_quali}
\end{figure*}

\begin{table*}
\begin{center}
\footnotesize
\begin{tabular}{rc|ccccc|ccccc}
\toprule
& & \multicolumn{5}{c|}{RealEstate10K-test} & \multicolumn{5}{c}{DL3DV10K} \\
\midrule
 Method & $\genloss$ & Rot.$\downarrow$ &  Trans.$\downarrow$ & RRA@5$\degree$$\uparrow$ &  RTA@5$\degree$$\uparrow$  & mAA@30$\degree$$\uparrow$ & Rot.$\downarrow$ &  Trans.$\downarrow$ & RRA@5$\degree$$\uparrow$ &  RTA@5$\degree$ $\uparrow$  & mAA@30$\degree$$\uparrow$ \\
\midrule  \midrule
pre-trained OS  & n.a. & 0.40 & 26.72 & 99.80\% & 11.63\% & 38.40\%  & 6.49 & 39.53 & 65.06\% & 2.81\% & 19.70\%\\
frozen OS + rec  & \xmark & \textbf{0.28} & \textbf{21.13} & \textbf{99.90\%} & \textbf{19.23\%} & \textbf{49.49\%} & \underline{4.54} & \underline{30.05} & \underline{76.27}\% & \textbf{7.88\%} & \underline{33.60\%}\\
\nameMethod & \cmark &  \underline{0.29}	& \underline{22.15} & \underline{99.80\%} & \underline{19.18\%} &	\underline{47.25\%} & \textbf{4.20}	& \textbf{29.17} &\textbf{78.63\%}& \underline{7.86\%} & \textbf{34.22\%} \\
\midrule
DUSt3R$^\dagger$ \cite{dust3r_cvpr24}  & n.a.  & 0.53 & 27.86 & 99.12\% & 13.70\% & 40.20\% & \textbf{1.33}& \underline{14.75} & \textbf{97.53\%} & \textbf{47.03\%} & \textbf{65.73\%} \\
DUSt3R*    & n.a. & \textbf{0.23} & \textbf{9.17} & \textbf{99.90\%} & \textbf{54.17\%} & \textbf {75.50\%} & 1.83 & \textbf{14.17} & \underline{93.63}\% & \underline{38.57\%} & \underline{64.40\%} \\
DUSt3R$^0$   & n.a. & \underline{0.32} & \underline{20.66} & \underline{99.70\%} & \underline{15.33\%} & \underline{47.47\%} & 6.25 & 39.37 & 65.84\% & 3.41\% & 19.73\%\\
 \bottomrule
 \end{tabular}
\end{center}
\vspace{-2mm}
  \caption{
 \textbf{V2C error comparison on RealEstate10K-test and DL3DV10K.} DUSt3R\textsuperscript{$\dagger$} indicates pre-trained \duster weights, whereas DUSt3R* is further trained with the same training set as our method -- RealEstate10K-train. DUSt3R$^0$ denotes training with RealEstate10K-train from scratch without initializing with pre-trained weights. In the sub-tables, \textbf{bold} is best in each sub-table; \underline{underlined} is the second place. \nameMethod produces good camera estimates on RealEstate10K-test and acceptable quality on out of distribution  data DL3DV10K. 
 }
 \label{tab:rec_quanti}
 \vspace{-1mm}
\end{table*}

\minisection{Metrics.}
We use the standard FID~\citep{fid} and FVD~\citep{unterthiner2019fvd} metrics to measure image and video quality, respectively. However, since such metrics do not reflect 3D consistency in the generated videos, we additionally adopt the recently proposed \meter\cite{asim24met3r} metric. We estimate the per-frame DINO~\cite{Caron21} features and warp them to the first frame using the estimated 3D point maps. We report the average cosine similarity between the features of the first frame and any other warped feature map weighted by the visibility masks.

We validate the quality of camera pose estimation on real videos (V2C) by comparing the estimated camera poses $(\textbf{R},\textbf{t})$ against the ground truth poses $(\bar{\textbf{R}},\bar{\textbf{t}})$. 
For rotation, we compute the relative error between two rotation matrices \cite{wang2023posediffusion}. 
Since the estimated and ground-truth translations can differ in scale, we follow \cite{wang2023posediffusion} to compute the angle between the two normalized translation vectors, \ie, $\arccos(\textbf{t}^\top\bar{\textbf{t}}/(\|\textbf{t}\|\|\bar{\textbf{t}}\|))$.
Besides reporting the average of the two errors, we also follow \cite{dust3r_cvpr24} to report Relative Rotation Accuracy (RRA) and Relative Translation Accuracy (RTA), \ie, the percentage of camera pairs with rotation/translation error below a threshold.
Due to the small number of frames handled by the video generator, each video sequence exhibits small rotation variation. Hence, we select a threshold 5$\degree$ to report RTA@5 and RRA@5. 
Additionally, we calculate the mean Average Accuracy (mAA@30), defined as the area under the curve accuracy of the angular differences at min(RRA@30, RTA@30).

\subsection{Generation Evaluation}
For each method, we generate 180 videos using the captions in RealEstate10K-test. We report the FID/FVD against  the real images/videos in RealEstate10K-test as well as the \meter metric where we use \nameMethod to estimate point maps. 
Table~\ref{tab:gen_quanti} suggests that our full model generates more realistic images/videos than pre-trained OpenSora (rows 1 and 4). Third row corresponds to a baseline where $\recloss$ is disabled by removing \duster decoders/heads, \ie, it is equivalent to standard diffusion model fine-tuning except only the weights of the temporal attention layers are updated. We see that fine-tuning with more data leads to lower FVD but not lower \meter error. In contrast,  \nameMethod achieves the lowest \meter error. This suggests that 3D consistency of video generators cannot simply be improved with more data whereas additional 3D-aware training tasks (point map estimation in our case) and losses are more effective. Finally, if we fine-tune the video model only with the reconstruction loss (row 2), while 3D consistency is improved, FID or FVD degrades. 
This is intuitive because without the generation loss, there is nothing to enforce the model to retain its full generation capability. 
The supplemental webpage shows examples of videos generated along with their \meter error maps.

\subsection{Reconstruction Evaluation}
In Table~\ref{tab:rec_quanti}, we compare the camera pose estimation (V2C) errors on RealEstate10K-test and the withheld DL3DV10K.
When we freeze the video model and only train the reconstruction decoder (row 1), we get noticeably worse results than the trainable counterparts (rows 2 and 3). This shows that the \textit{raw features from the pre-trained video generators are not fully 3D consistent} and hence ill suited for point map estimation. Furthermore, we see that the models trained with reconstruction loss only (row 2) lead to overall similar results to our full model, which uses both reconstruction and generation losses. This confirms that the two tasks are compatible and do not degrade each other's performance.

Our full method, \nameMethod, performs overall better than pre-trained DUSt3R$^\dagger$ (rows 3 vs 6) but is inferior than DUSt3R* (row 5) which fine-tunes DUSt3R$^\dagger$ with our data.
This suggests that the pre-trained DUSt3R weights, which were trained towards a 3D-aware task, contain richer information for camera pose estimation than the pre-trained OpenSora weights. In the future, we would like to explore training the video model from scratch using both generation and reconstruction losses, which we believe will result in a more 3D-consistent feature space.
Meanwhile, the overall results in Table \ref{tab:rec_quanti} suggests \nameMethod is not as competitive in generalization, which we attribute to the faster motion in DL3DV10K. DUSt3R$^\dagger$ and DUSt3R* contain pre-trained knowledge learned from matching wide-baseline view pairs. 

Figure~\ref{fig:rec-visuals} shows the qualitative comparison of our method and baselines in terms of point maps and resultant camera estimates. Since camera poses are estimated through registration, which builds 3D correspondences along the way, we visualize the final camera trajectories as well as the correspondence between the first and the last frame.
One can see that our method produces good camera trajectories similar to pre-trained \duster, which is a method tailored for reconstruction only, but without generation loss. In some cases ours are even closer to ground truth camera paths. The tests indicate that video generation and 3D-consistency are compatible tasks, and help improve each other. 


\minisection{Self consistency of T2V$\rightarrow$V2C and T2V+C}.
Since \nameMethod can \textit{generate} camera trajectories in two ways -- cascading T2V and V2C or the tightly coupled T2V+C pipeline -- it is worth comparing how much the two results differ.
To test this, we run the two pipelines with 100 prompts and report 0.45$\degree$ average difference in rotation and 19.20$\degree$ in translation; both are low errors compared with the corresponding numbers in Table \ref{tab:rec_quanti}, indicating that the camera poses from joint T2V+C pipeline are \textit{consistent} with T2V$\rightarrow$V2C.
The qualitative results in Figure~\ref{fig:t2vc_quali} also confirm this conclusion.

%% file: sec/05_limitation_conclusion.tex
\section{Conclusions and Future Work}
We have demonstrated that the tasks of video generation and pointMap estimation (equivalently, camera estimation for static scenes) can be made to be compatible and hence can be simultaneously trained with a unified architecture. To our knowledge, \nameMethod is the first method for joint video generation and 3D point map prediction. 
The new unified architecture enables end-to-end training of the two tasks, generates realistic and 3D-consistent videos, and achieves competitive camera pose estimation performance. This finding reveals that the tasks are synergetic -- video generators can be made 3D-consistent \textit{without} loss of visual quality, and its internal features reused for 3D-aware tasks. 

Since it is not trivial to obtain accurate camera annotations for dynamic scenes, our analysis is limited to videos of static scenes only. An important future direction is to analyze how video generation features adapt to camera estimation for dynamic scenes. Also, the length of the video sequences our method can handle is currently limited by the number of frames the generator can synthesize. 
As the video generators continue to improve to enable generation of longer sequences, our proposed solutions will also naturally extend to handling longer videos with larger baseline.


%% file: sec/10_suppmat.tex
The supplementary material consists of this document and the webpage. We provide qualitative results in our webpage.

\section{Temporal Regularizer Terms}
Given the estimated camera pose $(\textbf{R}_f,\textbf{t}_f)$ and focal length $l_f$ for each frame $f$, we define
$\flcloss$ and $\temploss$ as below:
\begin{align}
\flcloss &= \sum_f \left \| l_f - l_{f+1}\right \|_2^2 \\
\temploss &= \sum_f \left \| \textbf{t}_f - \textbf{t}_{f+1}\right \|_2^2 + \left \| \textbf{v}_f - \textbf{v}_{f+1}\right \|_2^2
\end{align}
where $\textbf{v}_f = \textbf{t}_f - \textbf{t}_{f+1}$. The first term of $\temploss$ encourages static camera position while the second term encourages constant velocity.

\section{Architectural Design Choices}
In Table 1 below, we compare the quality of estimated camera poses using the feature maps in different DiT block $b^i$.
Row 1c and 2c correspond respectively to our full \nameMethod model (row 1c in Table 1 and 2 in the main paper), which is $i$=26.
We first confirm using the features one block later, $i$=27, does not result in significant difference (row 1d vs.~1c; 2d vs.~2c). 
Next, we consider $i$=10 and 20, representing approximately one-third and two-third of total DiT blocks.
We see that compared to our results in the main paper (row 1c and 2c), $i$=10 yields lower errors in RealEstate10k-test but higher errors in DL3DV10k (row 1a and 2a), suggesting that using features of earlier blocks has a higher risk of poor generalization. Meanwhile, $i$=20 attains the lowest errors in both datasets, while ours remains on-par.
We therefore conclude that \textit{features of later blocks, e.g., $i\in [20, 27]$ is preferred than earlier blocks, and all later blocks should lead to similar results}. 
Instead of solely relying on one block $b^i$, one can potentially devise a module fusing features of all DiT blocks and projecting to the input space of \duster decoders, which we consider future work.

\begin{table*}[b]
\begin{center}
\footnotesize
\begin{tabular}{llrrrrr}
\toprule
 Backbone / Method & $b^i$  & Rot.~err.~($^\circ$) $\downarrow$ &  Transl.~err.~($\degree$) $\downarrow$ & RRA@5$\degree$ $\uparrow$ &  RTA@5$\degree$  $\uparrow$  & mAA@30$\degree$ $\uparrow$ \\
\midrule
& & & RealEstate10k \\
\midrule

1a.~trainable DiT  & $b^{10}$ & 0.28 & 18.94 & 99.94\% & 18.75\% & 52.63\% \\
1b.~trainable DiT  & $b^{20}$ & 0.28 & 19.10 & 99.85\% & 21.59\% & 53.35\% \\
1c.~trainable DiT & $b^{26}$ (\nameMethod) &  0.29	& 22.15 & 99.79\% & 19.18\% &	47.25\%  \\
1d.~trainable DiT    & $b^{27}$ & 0.28	& 23.09 & 99.80\% & 17.74\% &	46.88\%  \\
\midrule
& & & DL3DV10k \\
\midrule
2a.~trainable DiT  & $b^{10}$ & 4.68 & 30.90 & 73.87\% & 6.08\% & 30.76\% \\
2b.~trainable DiT  & $b^{20}$ & 4.01 & 27.71 & 77.91\% & 10.56\% & 36.62\%\\
2c.~trainable DiT & $b^{26}$ (\nameMethod) & 4.20	& 29.17 & 78.63\%& 7.86\% & 34.22\% \\
2d.~trainable DiT    & $b^{27}$ & 4.65	& 30.36 & 77.15\% & 7.98\% &	32.86\%  \\
 \bottomrule
 \end{tabular}
\end{center}
\vspace{-2mm}
  \caption{
 \textbf{Ablation study on which feature maps $b^i$ get passed to \duster.} Row 1c and 2c correspond respectively to our full \nameMethod model (row 1c in Table 1 and 2 in the main paper).
 }
 \label{tab:ablation} 
\end{table*}